\documentclass[10pt,twocolumn,letterpaper]{article}

\usepackage[pagenumbers]{wacv} 
\usepackage[accsupp]{axessibility}  
\usepackage{multirow}
\usepackage{arydshln}
\usepackage{comment}

\definecolor{wacvblue}{rgb}{0.21,0.49,0.74}
\usepackage[pagebackref,breaklinks,colorlinks,allcolors=wacvblue]{hyperref}

\def\wacvPaperID{673} 
\def\confName{WACV}
\def\confYear{2026}

\title{Action Anticipation at a Glimpse:\\ To What Extent Can Multimodal Cues Replace Video?}

\author{
Manuel Benavent-Lledo$^{1}$, 
Konstantinos Bacharidis$^{2,3}$, 
Victoria Manousaki$^{2,3}$,\\
Konstantinos Papoutsakis$^{2,4}$, 
Antonis Argyros$^{2,3}$, 
Jose Garcia-Rodriguez$^{1}$ \\
\normalsize
$^{1}$Universidad de Alicante, Spain
\normalsize
$^{2}$Foundation for Research and Technology-Hellas, Greece\\
\normalsize
$^{3}$University of Crete, Greece
\normalsize
$^{4}$Hellenic Mediterranean University, Greece\\
\normalsize
{\tt mbenavent@dtic.ua.es}
}

\begin{document}
\maketitle
\begin{abstract}
Anticipating actions before they occur is a core challenge in action understanding research. While conventional methods rely on extracting and aggregating temporal information from videos, as humans we can often predict upcoming actions by observing a single moment from a scene, when given sufficient context. Can a model achieve this competence? The short answer is yes, although its effectiveness depends on the complexity of the task. In this work, we investigate to what extent video aggregation can be replaced with alternative modalities. To this end, based on recent advances in visual feature extraction and language-based reasoning, we introduce AAG, a method for Action Anticipation at a Glimpse. AAG combines RGB features with depth cues from a single frame for enhanced spatial reasoning, and incorporates prior action information to provide long-term context. This context is obtained either through textual summaries from Vision-Language Models, or from predictions generated by a single-frame action recognizer. Our results demonstrate that multimodal single-frame action anticipation using AAG can perform competitively compared to both temporally aggregated video baselines and state-of-the-art methods across three instructional activity datasets: IKEA-ASM, Meccano, and Assembly101.



\end{abstract}    
\vspace{-0.6cm}
\section{Introduction}
\label{sec:intro}

The ability to anticipate human actions in videos enables a wide range of real-world applications, allowing systems to plan and respond in a proactive manner, 
rather than to act reactively. For instance, in autonomous driving, action anticipation enhances safety by predicting the behavior of distracted pedestrians or erratic drivers~\cite{rasouli2020pedestrianactionanticipationusing,liu2020spatiotemporal,girase2021loki}. In industrial settings, forecasting upcoming actions can help prevent defects and enhance workplace safety~\cite{Sener_2022_CVPR,Ben-Shabat_2021_WACV}. Similarly, accurately predicting human actions before they occur is essential for supporting fluency and success in human-robot collaborative scenarios~\cite{jang2020etri,hwang2021eldersim,dai2022toyota}.

Video understanding tasks, such as action anticipation, are computationally demanding due to the need for frame-wise feature extraction and temporal aggregation.
Transformer architectures have significantly contributed to long-range temporal modeling, addressing limitations of recurrent neural networks, which often struggle with preserving dependencies over extended sequences~\cite{lai2024human}. While recent research has explored different strategies to replace longer dependencies with a long short-term transformer~\cite{NEURIPS2021_08b255a5,zhao2022real}, or with textual information~\cite{Manousaki_2023_ICCV,benaventlledo2024enhancingactionrecognitionleveraging,zhao2023antgpt,liu2019roberta,bacharidis2021extracting}, the question remains: \emph{can short-term video aggregation be effectively substituted with alternative modalities?} 

\begin{figure}[t]
    \centering
    \includegraphics[width=\linewidth]{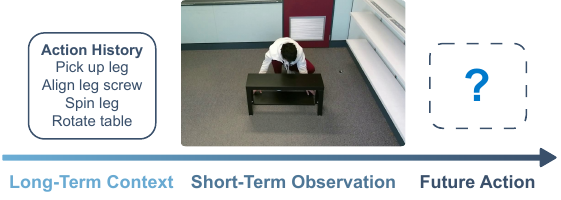}
    \vspace{-0.65cm}
    \caption{
    \textbf{How much information is necessary to anticipate the next action accurately?} Is a single-frame, short-term observation sufficient, or does long-term context from past actions improve prediction?
    We demonstrate that fusing a single frame with appropriate contextual modalities enables effective anticipation without relying on video aggregation.
    The frame is extracted from the IKEA-ASM~\cite{Ben-Shabat_2021_WACV} dataset, and the next action is \emph{attach shelf to table}.
    }
    \label{fig:Concept_fig}
    \vspace{-0.6cm}
\end{figure}

Future is inherently uncertain, therefore, as humans we might or might not be able to determine the upcoming actions given a frame or a video sequence (\cref{fig:Concept_fig}). Some works directly tackle the uncertainty problem or acknowledge various possible future events~\cite{Suris_2021_CVPR,10.1007/978-3-319-46478-7_51}. Nonetheless, having more information allows for more accurate predictions.
Self-supervised methods capture intrinsic features of the data~\cite{oquab2023dinov2,Caron_2021_ICCV,10.1145/3577925}. Language models, including Vision-Language Models (VLMs)~\cite{sanh2019distilbert,dubey2024llama,wu2024deepseek,devlin2019bert,openai2024gpt4technicalreport}, provide additional knowledge that can inform predictions. Together, these approaches help extract the most relevant cues about past events from a single image.


In this work, we investigate to what extent a single frame, enriched with multimodal information, can provide sufficient context for effective action anticipation.
To this end, we introduce AAG, a method for Action Anticipation at a Glimpse. Our approach uses RGB data as the primary input modality given its broad accessibility.
To complement RGB and enhance spatial understanding, we propose incorporating additional modalities. While prior work~\cite{RAGUSA2023103764,Zhang_2024_WACV,Feng_2025_CVPR} has demonstrated the value of fine-grained visual features (e.g., hands, objects, or pose) these often depend on fine-tuned, task-specific detectors, and are more susceptible to occlusion or perspective bias, limiting their applicability. Instead, depth provides robust spatial cues that are consistent across viewpoints, easy to obtain, and independent of fine-tuned models. In addition to RGB and depth features, we incorporate semantic priors derived from previously observed actions to provide long-term context.

To obtain semantic priors, AAG uses language models to encode descriptive summaries of preceding actions. These are generated either from the predictions of an action recognition model, or inferred from the current frame using VLMs. In our experimental analysis, we assess the contribution of each modality using the most effective fusion strategy. Additionally, we conduct a comprehensive ablation study to examine how action history is generated and represented, analyzing both its source and temporal extent.

Our main contributions can be summarized as follows:
\begin{itemize}
    \item We investigate the potential of single-frame action anticipation as a competitive alternative to video-based approaches by leveraging recent advances in multimodal learning, vision models, and language models.
    \item We introduce AAG, a method for action anticipation at a glimpse, and evaluate it through extensive experiments on modality contributions, prior actions, and comparisons with video-based methods. Code available on GitHub\footnote{\url{https://github.com/ManuBenavent/AAG}}.

    
    \item We demonstrate that multimodal single-frame anticipation competes with state-of-the-art video-based methods on three diverse datasets (IKEA-ASM~\cite{Ben-Shabat_2021_WACV}, Meccano~\cite{RAGUSA2023103764}, Assembly101~\cite{Sener_2022_CVPR}), offering insights into when temporal reasoning or specific modalities are most beneficial.
\end{itemize}

\section{Related Work}
\label{sec:related}
In this section, we review multimodal video-based action anticipation methods and then focus on single-frame action recognition and anticipation methods, given the limited work in this area.

\noindent \textbf{Video-based Human Action Anticipation} has been widely explored in recent years~\cite{lai2024human}. Some methods rely solely on RGB information, achieving state-of-the-art results through strong temporal aggregation mechanisms. One such example is TempAgg~\cite{TempAgg}, which presents a flexible multi-granular temporal aggregation framework to address the challenges of long-range video understanding. 
The Anticipative Video Transformer (AVT)~\cite{AVT} is an end-to-end attention-based model that jointly anticipates actions and learns frame encoders predictive of future content. By maintaining the sequence of observed actions, AVT captures long-range temporal dependencies effectively.


In contrast, multimodal methods incorporate features from multiple sources to build richer representations before classification. For instance, Ragusa et al.~\cite{RULSTM} introduce RULSTM, a Rolling-Unrolling LSTM for action anticipation, with a Rolling LSTM encoding observed frames to capture temporal dependencies and an Unrolling LSTM anticipating future actions and object interactions. Textual information, typically in the form of action descriptions, has also proven valuable for improving anticipation performance. The Visual-Linguistic Modeling of Action History framework~(VLMAH)~\cite{Manousaki_2023_ICCV} integrates immediate visual features with a lightweight linguistic representation of past actions, capturing both recent and distant context. AntGPT~\cite{zhao2023antgpt} uses large language models to extract goal-directed activities, also referred to as high-level actions, which improve fine-grained action prediction. A similar strategy is used in~\cite{zatsarynna2023action}, where a joint loss enforces consistency between fine- and coarse-grained actions.

\noindent \textbf{Single-Frame Human Action Recognition} is inherently challenging due to the absence of temporal cues. To address this, various methods leverage knowledge distillation, contextual reasoning, and enhanced feature extraction. The work in~\cite{gkioxari2015contextual} emphasizes the importance of jointly modeling actors and their environment using region-based CNNs, demonstrating how context significantly improves recognition performance.
Recent approaches to single-frame action recognition focus on spatial reasoning and feature learning. Ashrafi et al.~\cite{ashrafi2023still} model relationships between body joints and objects to capture human–object interactions. Liang et al.~\cite{liang2024patch} propose a patch-based method that localizes actions in still images without bounding boxes. Further improvements were introduced through transformer-based models and knowledge distillation. In~\cite{hosseyni2024human} vision transformers are used to extract spatial relations between poses and scene context, and Saleknia and Ayatollahi~\cite{saleknia2024multi} improve model efficiency via progressive knowledge distillation. Other methods such as~\cite{he2023context} enhance contextual reasoning in still images using pixel- and region-level attention. 

\noindent \textbf{Single-Frame Human Action Anticipation} has received limited attention despite its computational advantages, such as removing the need for temporal aggregation and improving spatial feature learning when combined with video information~\cite{lei2022revealing}. Due to the limited context in a single frame, existing approaches are mostly multimodal. For example, Lan et al.~\cite{lan2014hierarchical} propose hierarchical ``movemes'', a multi-level representation capturing human movement at varying granularities for anticipating future actions from a single image or short clip. Vondrick et al.~\cite{vondrick2016anticipating} introduce a self-supervised model that predicts high-level visual features of future frames from a single image, generating diverse outcomes to handle uncertainty.
VisualCOMET~\cite{park2020visualcomet} generates dynamic context from still images by reasoning about past events, intentions, and future outcomes, offering high-level commonsense understanding rather than explicit action prediction. Similarly, ActionCOMET~\cite{sampat2024actioncomet} infers commonsense knowledge related to actions in images, including preconditions, outcomes, intentions, and sequences, by leveraging language models and annotated cooking videos. Although neither VisualCOMET nor ActionCOMET perform direct action anticipation, they provide valuable semantic insights to support downstream reasoning tasks.

\begin{figure*}[!thbp]
\centering
\includegraphics[width=0.8\textwidth]{figures/pipeline.pdf}
\caption{AAG architecture with proposed action history strategies. Given a frame $x_T$, captured $\delta$ seconds before an action, the visual fusion module combines RGB and depth embeddings from a frozen extractor (depth mapped to RGB for encoding). Visual features are fused with past actions via a self-attention transformer. We test three history encoding methods: (1) prompting a vision-language model (blue), (2) generating past action descriptions (red), and (3) separately encoding action classes before fusion (yellow), which performs best.}
\label{fig:method}
\vspace{-0.3cm}
\end{figure*}


\section{AAG: Action Anticipation at a Glimpse}
\label{sec:method}

We introduce AAG, depicted in \cref{fig:method}, to assess action anticipation performance using minimal short-term visual observations, and to evaluate the effectiveness of multimodal information as an alternative to video-based methods.



\subsection{Visual Encoder}
Given a video sequence, part of a longer video up to time $t$, consisting of $N$ frames $X_i = {x_1, \dots, x_N}$. The goal of action anticipation is to predict the upcoming actions at $t + \delta$, where $t$ is the current instant and $\delta$ is the anticipation time. Video-based approaches typically define a temporal window of size $w$, using frames from $x_{N-w}$ to $x_N$ to extract a spatio-temporal representation. However, the choice of $w$ is constrained by computational limits, which restrict the ability to capture long-range temporal dependencies.

Instead, we argue that a single frame, $x_N$, can provide sufficient information when fused with the appropriate modalities. To this end, rather than extracting features from CNN-based backbones~\cite{wang2018temporal,szegedy2016rethinking,he2016deep}, a more refined representation is required. Transformer architectures, and particularly self-supervised methods, offer a suitable solution for this problem due to their ability to extract the underlying representations of data. To this end, we select DINOv2~\cite{oquab2023dinov2} as the primary backbone for our method, as it does not require fine-tuning to generate high-quality representations, producing the best features compared to vision transformers~\cite{dosovitskiy2020image,radford2021learning} and CNN-based methods~\cite{wang2018temporal}. We denote the representation from DINOv2 CLS token as $m_{RGB} \in \mathbb{R}^{1 \times D_{ft}}$, where $D_{ft}$ is the embedding dimension.

RGB data is highly effective for capturing relationships between objects and actors, but it lacks spatial cues necessary for distinguishing foreground elements like objects or people. As previously remarked, additional or RGB-derived modalities (e.g., objects, hands, or pose) can complement RGB cues by providing additional insights, though they often depend on specific viewpoints and/or fine-tuned detectors. 
In contrast, depth captures the distance between the camera and scene elements, providing explicit spatial information. Besides, it is an accessible alternative due to its ease of acquisition, the effectiveness of recent depth estimation techniques, and its robustness across varying camera views.

The contribution of depth, however, can sometimes be misleading due to noise (see Appendix A for detailed qualitative and quantitative results on the depth modality). Additionally, unlike RGB images, depth frames are single-channel, limiting the use of standard feature extractors.
To address these limitations, we use Depth Anything V2~\cite{NEURIPS2024_26cfdcd8} to generate a high quality depth estimation, mitigating both the absence of depth information in some datasets, and the quality of depth frames. Similar to the depth maps provided in~\cite{RAGUSA2023103764}, depth frames are converted into RGB representations by applying color mapping. This transformation enables the use of standard feature extractors, originally designed for RGB images such as DINOv2. As a result, we obtain a depth feature vector denoted as $m_{Depth} \in \mathbb{R}^{1 \times D_{ft}}$.

To effectively integrate both feature spaces into a single representation that combines RGB and depth cues, we employ a cross-attention transformer encoder. This enables each modality to mutually enhance the other’s representation, resulting in a richer, more comprehensive embedding. We compute cross-attention as follows:
\vspace{-0.1cm}
\begin{equation}
    Q = W_Q m_{\text{RGB}}, \quad K = W_K m_{\text{Depth}}, \quad V = W_V m_{\text{Depth}}
    \label{eq:ca-vectors}
\end{equation}
\vspace{-0.3cm}
\begin{equation}
    m_{\text{vis}} = \text{softmax} \left( \frac{Q K^T}{\sqrt{D}} \right) V,
    \label{eq:attention}
\end{equation}

where \( W_Q, W_K, W_V \in \mathbb{R}^{D \times D} \) are learnable projection matrices that map the input embeddings into query (\(Q\)), key (\(K\)), and value (\(V\)) representations. The resulting representation is \( m_{\text{vis}} \in \mathbb{R}^{1 \times D} \), where \( D \) is the embedding dimension of AAG. To project the feature embeddings of size $D_{ft}$ into AAG's feature space a linear layer is used.

Although AAG is designed to operate on single-frame inputs, its modular visual encoder architecture supports both single-frame and video-based inputs. To explore this flexibility, we use a standard video transformer from prior work~\cite{benaventlledo2024enhancingactionrecognitionleveraging,Wang_2021_ICCV} to assess the temporal aggregation effect. This approach demonstrates superior performance compared to self-supervised video-based methods~\cite{bertasius2021space,assran2025v}, as evidenced in Appendix A. The video encoder independently aggregates frame-level RGB and depth features, which are then fused using the standard AAG cross-attention mechanism.

\subsection{Action History Encoder}
The future is inherently uncertain, and past information can help disambiguate upcoming actions. Prior work has shown that longer temporal dependencies can be effectively modeled using textual formats~\cite{Manousaki_2023_ICCV,benaventlledo2024enhancingactionrecognitionleveraging}. We explore three different strategies to model previous actions for AAG based on how past actions are obtained and how they are aggregated.

\noindent \textbf{VLM Prompting:}
As humans, we can often infer a set of past actions based on contextual information and a single image. This ability is particularly useful in constrained environments, such as industrial settings, where the sequence of actions exhibits less variability than in open-world tasks.

Recently, VLMs have shown strong reasoning abilities, surpassing captioning and VQA models like BLIP-2~\cite{li2023blip}, which generate descriptions solely from visual features, ignoring prompt context (see Appendix B for comparison). We evaluate three VLMs (Llama-3.2 Vision (11B)~\cite{dubey2024llama}, DeepSeek-VL2 Tiny~\cite{wu2024deepseek}, and GPT-4o mini~\cite{openai2024gpt4technicalreport}) under prompts with varying contextual detail, from no context (e.g., \emph{``What are the previous actions that led up to this frame?''}) to refined prompts specifying dataset domain (e.g., \emph{assembling Ikea furniture}), number of past actions, and possible action classes.

We encode the resulting description using a language model, leveraging its ability to model long text sequences. In particular, DistilBERT~\cite{sanh2019distilbert} achieves the best performance for this task compared to similar text encoders~\cite{radford2021learning,devlin2019bert,liu2019roberta}. The output representation extracted from the CLS token is denoted as $m_{txt} \in \mathbb{R}^{1 \times D_{txt}}$, where $D_{txt}$ is the embedding size of the selected text encoder. 

\noindent \textbf{Seen Actions Encoding:}
Alternatively, instead of relying on a VLM, we leverage previous observations to automatically generate similar action history descriptions. This approach not only reduces the computational overhead associated with vision-language models, but also improves the accuracy and consistency of the past action history by grounding it in observed frames rather than single-frame reasoning.

To this end, we use $N$ prior actions to generate descriptions, where $N$ varies across datasets. However, our experiments demonstrate that using 3 to 7 previous actions yields the best results. During training, prior actions are obtained from ground truth. At inference time, in a realistic setting, we reuse the anticipation features for action recognition, thus avoiding additional frame processing. As in the VLM-base approach, the action sequence in text form is encoded with a language model, yielding $m_{txt} \in \mathbb{R}^{1 \times D_{txt}}$.

\noindent \textbf{Seen Actions Aggregation:} Finally, we propose encoding each action class individually and aggregating their textual representations, eliminating the need for a textual encoder at inference time, since the set of classes is known beforehand. AAG uses concatenation of these per-action embeddings as the aggregation mechanism, which is both more computationally efficient and yields the best performance. The resulting representation is denoted as $m_{txt} \in \mathbb{R}^{1 \times (N \cdot D_{txt})}$.

\subsection{Multimodal Fusion \& Classification}
Visual and textual cues are finally aggregated using a self-attention transformer encoder. Before fusion, we project $m_{vis}$ and $m_{txt}$ to ${D}$, the embedding dimension of AAG, ensuring that both modalities lie in $\mathbb{R}^{1 \times D}$ to avoid modality imbalance. The joint input sequence is formalized as:
\begin{equation}
    M = [m_{vis};m_{txt}] \in \mathbb{R}^D.
\end{equation}
Self-attention vectors are calculated as:
\begin{equation}
    Q = W_Q M, \quad K = W_K M, \quad V = W_V M,
\end{equation}
where $W_X$ are learnable matrices as in \cref{eq:ca-vectors}, and the attention operation is computed as in \cref{eq:attention}. The resulting embeddings $m_{fus} \in \mathbb{R}^{2 \times D}$ are averaged to produce a final embedding in $\mathbb{R}^D$, which is used for classification. A linear classifier is then applied to this embedding, followed by a softmax layer to predict class probabilities. Cross-entropy loss is used for supervision during training.

\section{Experimental Setup}
\label{sec:setup}

\subsection{Datasets}
We evaluate our approach on three publicly available datasets designed for action recognition and anticipation in industrial scenarios. They encompass diverse environments, action granularity levels, and procedural variability.
 
\textbf{IKEA-ASM}~\cite{Ben-Shabat_2021_WACV} contains 371 third-person videos of furniture assembly across five environments, featuring 33 atomic actions and 4 high-level activities. RGB and depth data are captured from three different views, we use the top RGB view to align with the depth perspective. The default environment-based train/test split (254/117 videos) is used.

\textbf{Meccano}~\cite{RAGUSA2023103764} is an egocentric dataset comprising 20 videos of participants assembling a toy motorbike, recorded with head-mounted RGB and depth sensors. It includes 61 action classes derived from 20 object and 12 verb categories. We use the standard 11/9 train/test split.

\textbf{Assembly101}~\cite{Sener_2022_CVPR} contains 362 videos of participants assembling and disassembling 101 toy vehicles without fixed instructions, resulting in diverse execution patterns. It includes over 1M action segments spanning 1380 fine-grained action classes, derived from 90 objects and 24 verbs, captured across 12 camera views (8 RGB close-up and 4 monochrome egocentric). We use the exocentric (third-person) camera \emph{v4}, which provides the best RGB performance~\cite{Sener_2022_CVPR}, and follow the train/validation splits from~\cite{Manousaki_2023_ICCV}, as the official test set is not publicly available.

\subsection{Methods \& Implementation Details}

\noindent \textbf{Evaluation:} We evaluate our method by predicting actions $\delta = 1s$ into the future and report performance using two standard metrics in action anticipation: top-$k$ accuracy and class-mean Top-5 recall~\cite{Manousaki_2023_ICCV,RAGUSA2023103764,Sener_2022_CVPR,damen2022rescaling}. Top-$k$ accuracy captures uncertainty in future predictions, while class-mean Top-5 recall addresses long-tail class distributions. Following prior work, we adopt top-$k$ accuracy as the primary metric for IKEA-ASM and Meccano~\cite{Manousaki_2023_ICCV,RAGUSA2023103764}, and class-mean Top-5 recall for Assembly101~\cite{Sener_2022_CVPR}. Bold indicates the best performance for the respective primary metric in tables.

To evaluate the role of action history, we report results using both ground-truth and predicted histories. As previously noted, predictions are generated by an action recognition model that  shares the same architecture and input modalities as our anticipation setup, ensuring consistency.

As established in action recognition benchmarks~\cite{Ben-Shabat_2021_WACV,RAGUSA2023103764,Sener_2022_CVPR}, we use top-1 accuracy as evaluation metric. While full results are provided in the supplementary material, we report key findings here. Using an RGB frame and action history, the model achieves 66.43\% on IKEA-ASM, 31.21\% on Meccano, and 34.19\% on Assembly101. Adding depth yields: 66.84\%, 31.14\%, and 32.77\%, respectively.

\noindent \textbf{Implementation Details:} We adopt DINOv2~\cite{oquab2023dinov2} and DistilBERT~\cite{sanh2019distilbert} as the visual and text encoders, respectively. Depth frames are estimated using Depth Anything V2~\cite{NEURIPS2024_26cfdcd8}. Transformer encoders are configured with 2 layers, 4 attention heads per layer, and an embedding dimension $D$ set to 768. 
As described in Section~\ref{sec:method}, AAG employs a cross-attention transformer for visual feature fusion, followed by a self-attention transformer for multimodal integration. An extended ablation on encoder configurations, fusion strategies, and depth sources is provided in Appendix A.

Video-based experiments on AAG use a dedicated temporal encoder to aggregate precomputed frame-level features. This encoder follows the best-performing configuration from prior work~\cite{benaventlledo2024enhancingactionrecognitionleveraging,Wang_2021_ICCV}, and is implemented as a transformer with 3 layers, 8 attention heads per layer, sinusoidal positional encodings, and a CLS token.

Experiments involving action history (AH) use ground-truth sequences at training and predicted sequences at inference. Predictions are generated by an auxiliary action recognizer that shares the same architecture and inputs, trained with an action recognition objective. This design ensures consistency in the single-frame analysis of prior actions.

\noindent \textbf{Training Details:} AAG has been implemented using PyTorch, and experiments are conducted on Nvidia RTX 4090 GPUs. We employ AdamW optimizer with a weight decay of $0.01$ and a base learning rate of $5 \times 10^{-5}$. Models are trained for $100$ epochs, with an early stopping patience of $10$ epochs, and an improvement threshold of $0.001$. Batch size is set to 32. Video-based experiments on AAG use the same training settings with an input window of 16 frames.

\noindent \textbf{SOTA Methods Details:} We compare AAG with four state-of-the-art video-based action anticipation methods that report results on the same benchmark datasets: AVT~\cite{AVT}, RULSTM~\cite{RULSTM}, TempAgg~\cite{TempAgg}, and VLMAH~\cite{Manousaki_2023_ICCV}. To ensure completeness and comparability across datasets, we re-train missing baselines using our evaluation protocol with consistent train/validation splits and camera views. For fairness in action history settings, the history length in VLMAH is constrained to match that of AAG. On AVT, we use pre-extracted features from the BNInception~\cite{ioffe2015batch} backbone of TSN~\cite{wang2018temporal}, while on Assembly101, TempAgg and VLMAH are adapted to the same camera views as AAG.

\section{Experimental Results}
\label{sec:results}

\begin{table*}[t]
    \centering
    \footnotesize
    \setlength{\tabcolsep}{3pt}
    \begin{tabular}{ll rr rr rr}
        \toprule
        \textbf{AH Source} & \textbf{Modalities}
 & \multicolumn{2}{c}{\textbf{IKEA-ASM}} & \multicolumn{2}{c}{\textbf{Meccano}} & \multicolumn{2}{c}{\textbf{Assembly101}} \\
        \cmidrule(lr){3-4} \cmidrule(lr){5-6} \cmidrule(lr){7-8}
        & & Top-1/5 Acc & Recall@5 & Top-1/5 Acc & Recall@5 & Top-1/5 Acc & Recall@5 \\
        
        \midrule
        \multirow{2}{*}{None} 
            & RGB  & 34.37 / 83.43 & 43.71 & 27.21 / 50.02 & 8.34 & 1.94 / 7.44 & 8.00 \\
            & RGB, Depth & 38.82 / 86.19 & 46.52 & \cellcolor{green!15}27.21 / 51.15 & 8.33 & 1.82 / 6.76 & 6.16 \\

        \midrule
        \multirow{3}{*}{GT} 
            & AH & \textbf{63.55 / 94.52} & 65.09 & 31.75 / 73.57 & 37.33 & 31.19 / 59.70 & \textbf{38.49} \\
            & RGB, AH & 60.14 / 91.00 & 53.28 & 30.01 / 66.44 & 18.98 & 27.37 / 52.88 & 26.12 \\
            & RGB, Depth, AH & 61.83 / 89.64 & 51.37 & 31.86 / 66.80 & 19.16 & 26.82 / 53.05 & 24.86 \\
            & RGB$_{vid}$, Depth$_{vid}$, AH & 63.15 / 91.76 & 60.97 & \textbf{33.88 / 67.90 }& 25.95 & 30.27 / 56.51 & 33.05 \\

        \midrule
        \multirow{4}{*}{Pred} 
            & AH* & 41.42 / 79.03 & 46.50 & 25.72 / 51.97 & 12.69 & 13.46 / 32.87 & \cellcolor{green!15}9.80 \\
            & RGB, AH & 43.82 / 83.11 & 47.26 & 26.22 / 52.18 & 13.63 & 13.72 / 30.93 & 8.25 \\
            & RGB, Depth, AH & \cellcolor{green!15}44.66 / 82.87 & 46.59 & 26.43 / 54.95 & 12.27 & 13.56 / 32.13 & 8.90 \\
            & RGB$_{vid}$, Depth$_{vid}$, AH & \underline{46.02 / 83.43} & 50.79 & \underline{27.24 / 54.52} & 12.53 & 14.01 / 31.97 & \underline{11.24} \\

        \bottomrule
    \end{tabular}
    \caption{Analysis of (a) the influence of modality selection on AAG performance across datasets, and (b) the effect of using ground-truth action history (GT) versus predicted (Pred). Underlined values and green-highlighted cells indicate the overall and single-frame best results, respectively, under realistic settings. * indicates AH is populated from AAG's action recognizer using RGB, Depth, AH modalities.}
    \label{tab:modalities}
    \vspace{-0.3cm}
\end{table*}


We conduct an experimental study on key factors influencing the performance of multimodal single-frame action anticipation, using AAG, and analyze its performance gap relative to top-performing video-based approaches.

\subsection{Impact of Multimodal Cues and Temporal Aggregation Under Ideal \& Realistic Settings}
Table~\ref{tab:modalities} shows the contribution of multimodal inputs (RGB, depth, action history) to single-frame action anticipation and the impact of temporal aggregation via a transformer encoder, enabling controlled comparison with video-based models under the same conditions. The single-frame approach remains competitive in both oracle (GT) and realistic (Pred) action history settings across datasets.

Among the individual modalities, self-supervised RGB features consistently show strong performance across datasets. Depth, however, exhibits more variable utility depending on the camera perspective. Specifically, it proves more beneficial in third-person views, as seen in the IKEA-ASM dataset (+4.45\%), where the exocentric perspective allows depth cues to effectively capture object localization and spatial relationships. In contrast, in datasets like Meccano or Assembly101, which rely on egocentric or top-down views, depth tends to be less informative, or even misleading. We attribute this to the constrained nature of these perspectives, which capture interactions within a narrow, nearly planar workspace with limited depth variation. Consequently, close-up datasets may benefit more from fine-grained visual cues related to hands, objects, and their interactions. Additional results on depth features, including qualitative results per dataset, are provided in Appendix A.

\begin{table*}[t]
    \centering
     \footnotesize
    \setlength{\tabcolsep}{3pt} 
    \begin{tabular}{llrr rr rr}
        \toprule
        \textbf{Method} & \textbf{Modalities} & \multicolumn{2}{c}{\textbf{IKEA-ASM}} & \multicolumn{2}{c}{\textbf{Meccano}} & \multicolumn{2}{c}{\textbf{Assembly101}} \\
        \cmidrule(lr){3-4} \cmidrule(lr){5-6} \cmidrule(lr){7-8}
        & & Top-1/5 Acc & Recall@5 & Top-1/5 Acc & Recall@5 & Top-1/5 Acc & Recall@5 \\
        \midrule
        AVT~\cite{AVT} & RGB$_{vid}$ & 27.10 / 69.70 & 45.78 & \underline{27.43 / 53.38} & 32.05 & 7.25 / 23.80 & \underline{17.03} \\
        TempAgg~\cite{TempAgg} & RGB$_{vid}$ & 26.90 / 70.14 & 16.42 & 19.69 / 25.37 & 17.47 & 11.43 / 34.16 & 9.07 \\
        RULSTM~\cite{RULSTM} & RGB$_{vid}$ & 26.37 / 70.05 & 16.00 & 24.08 / 58.23 & 22.38 & 5.95 /21.28 & 1.00 \\
        \textbf{AAG} & RGB & \underline{34.37 / 83.43} & 43.71 & 27.21 / 50.02 & 8.34 & 1.94 / 7.44 & 8.00 \\
        \midrule
        VLMAH~\cite{Manousaki_2023_ICCV} & RGB$_{vid}$, AH & \textbf{52.31 / 85.44} & 47.45 & \textbf{29.11 / 57.05} & 57.05 & 9.17 / 27.63 & \textbf{25.13} \\
        \textbf{AAG} & RGB, Depth, AH & 44.66 / 82.87 & 46.59 & 26.43 / 54.95 & 12.27 & 13.56 / 32.13 & 8.90  \\
        \bottomrule
    \end{tabular}
    \caption{Comparison of AAG with state-of-the-art video-based action anticipation methods. Action History (AH) is predicted by an action recognizer (single-frame for AAG, video-based for VLMAH). 
    Underline denotes the best RGB-only performance.
    }
    \label{tab:performance}
    \vspace{-0.3cm}
\end{table*}

Beyond visual features, incorporating a text-based embedding of recently performed actions leads to a substantial performance boost when combined with visual inputs, in both oracle (GT) and realistic (Pred) scenarios. This improvement arises from the temporal context that action history provides, enabling the model to disambiguate visually similar frames corresponding to different stages of execution. This effect is especially pronounced in datasets containing structured activities (more than one), that share common action steps, as in the case of IKEA-ASM and Assembly101 datasets. In such cases, past actions serve as a strong prior for anticipating what comes next, sometimes to the extent that reliance on visual input becomes less critical. Notably, we observe that using a perfect action history (GT) alone performs better than when combined with short-term observations obtained from single-frame or video sources in these settings. This suggests that action history itself encodes highly informative temporal cues. Under realistic settings, short-term observations contribute to more robust performance in both datasets, though with notable differences. IKEA-ASM achieves the best single-frame performance when visual modalities are fused with action history. In contrast, Assembly101 performs best with action history alone in the single-frame setup. This is attributed to the complexity of tasks in Assembly101, where more nuanced temporal modeling of short-term observations is required, which is effectively captured when video inputs are used. Interestingly, while visual inputs introduced noise in the performance of the ground-truth action history. Under realistic conditions, their contribution becomes two-fold: (1) generating the action history with an equivalent action recognizer and (2) refining the predictions of the action history by providing a more robust representation.

In contrast, the Meccano dataset shows weaker performance when using action history under realistic conditions, performing worse than the visual-only baseline. We attribute this to the lower accuracy of the action recognizer in this domain, which limits the quality of the predicted history, which contributed to a stronger performance under ground-truth action history. This highlights the crucial role of action recognition quality in the overall AAG performance when relying on predicted action histories. To better understand this dependency, we explore the impact of replacing the single-frame action recognizer with more advanced video-based architectures~\cite{Feichtenhofer_2019_ICCV,carreira2017quo,lin2019tsm}. Detailed results and comparative analyses are presented in Appendix A, along with further experiments on the effect of multimodality in video-based settings.

\subsection{Impact of Temporal Resolution on Action Anticipation: Single Frame vs. Video}

Table~\ref{tab:performance} presents a comparative analysis of AAG against leading video-based action anticipation methods across the selected benchmarks. On IKEA-ASM, AAG surpasses RGB-only baselines and performs on par with VLMAH when long-term context (AH) is included. This is enabled in part by self-supervised DINOv2 features, which provide rich spatial representations, and by IKEA-ASM’s constrained action space, where predictable task flows make single-frame cues sufficient for accurate anticipation, reducing the need for explicit temporal modeling.



A similar trend appears in Meccano, where AAG (RGB-only) matches or slightly outperforms several video-based baselines in Top-1 Acc., though it underperforms in Recall@5 due to limited temporal context. Meccano’s larger action set, finer labels, and long-tail distribution challenge recall-based metrics. Nonetheless, its linear task structure allows AAG to remain competitive by leveraging DINOv2’s spatial features and contextual cues from action history, compensating for the absence of temporal aggregation.


In contrast, Assembly101 highlights the limits of single-frame models in more complex and flexible environments. Activities can be performed in multiple valid orders with substantial variation in timing, making long-term temporal dependencies critical for accurate anticipation. Without sequential context, single-frame models struggle to distinguish visually similar frames representing different execution stages or functional roles. Video-based models like TempAgg address this by capturing both short- and long-term dependencies, while VLMAH benefits from strong recognition performance to generate action history. Thus, robust performance in dynamic, ambiguous scenarios like Assembly101 requires explicit temporal modeling.


Broadening the perspective beyond industrial domains, datasets vary in how closely their activities follow structured procedures.
When actions are predictable, models that anticipate future steps from a single frame are expected to perform well by relying on strong visual cues, often matching the performance of video-based models.
However, as the variability in procedures increases, models benefit more from longer temporal context, emphasizing the influence of dataset structure on performance.
Importantly, choosing the right model isn't just about accuracy, it also depends on the computational efficiency needed for real-world deployment.

\begin{table}[t]
    \footnotesize
    \centering
    \setlength{\tabcolsep}{3pt} 
    \begin{tabular}{lccc}
        \toprule
        \textbf{Method} & \textbf{\begin{tabular}[c]{@{}c@{}}Total \\ Params (M)\end{tabular}} & \textbf{\begin{tabular}[c]{@{}c@{}}Trainable\\  Params (M)\end{tabular}} & \textbf{\begin{tabular}[c]{@{}c@{}}\# Processed\\ Frames\end{tabular}} \\
        \midrule
        AVT~\cite{AVT} & 404 & 392 & 10 \\
        TempAgg~\cite{TempAgg} & 135 & 123 & 37 \\
        RULSTM~\cite{RULSTM} & 79 & 67 & 14 \\
        VLMAH~\cite{Manousaki_2023_ICCV} & \textbf{52} & 45 & 8 \\
        \textbf{AAG} & 202 & \textbf{24} & \textbf{1} \\
        \bottomrule
    \end{tabular}
    \caption{Computational cost and model size across methods.}
    \label{tab:compute_comparison}
    \vspace{-0.4cm}
\end{table}
    
\begin{table*}[t]
    \begin{minipage}{0.55\textwidth}
     \footnotesize
    \setlength{\tabcolsep}{3pt}
    \centering
    \begin{tabular}{lrr rr rr}
        \toprule
        \textbf{\begin{tabular}[c]{@{}l@{}}\# Past\\Actions\end{tabular}} & \multicolumn{2}{c}{\textbf{IKEA-ASM}} & \multicolumn{2}{c}{\textbf{Meccano}} & \multicolumn{2}{c}{\textbf{Assembly101}} \\
        \cmidrule(lr){2-3} \cmidrule(lr){4-5} \cmidrule(lr){6-7}
        & Top1/5 Acc & Rec@5 & Top-1/5 Acc & Rec@5 & Top-1/5 Acc & Rec@5 \\
        \midrule
        1 & 53.70 / 89.84 & 53.18 & 30.40 / 63.43 & 16.72 & 23.24 / 49.07 & 23.46 \\
        3 & 60.38 / 92.44 & 62.95 & \textbf{31.86 / 66.80}  & 19.16 & 26.82 / 53.05 & \textbf{24.86} \\
        5 & 57.70 / 90.48 & 60.63 & 29.51 / 62.82 & 15.98 & 25.47 / 51.06 &  23.89 \\
        7 & \textbf{61.83 / 89.64}  & 51.37 & 31.71 / 64.63 & 16.84 & 24.35 / 48.97 & 23.63 \\
        10 & 58.14 / 90.32 & 62.04 & 28.56 / 62.08 & 16.14 & 24.04 / 48.29 & 20.99 \\
        
        \bottomrule
    \end{tabular}
    \caption{Impact of the size of the past action history (memory pool) on AAG's performance for all examined datasets.}
    \label{tab:past_act_history}
\end{minipage}
\hfill
\begin{minipage}{0.4\textwidth}
\setlength{\tabcolsep}{3pt}
\footnotesize
\centering
\begin{tabular}{llr}
\toprule
\multicolumn{2}{c}{\textbf{Action History Source}} & \textbf{Top-1 / 5 Acc} \\\midrule
\multirow{3}{*}{\begin{tabular}[c]{@{}l@{}}Llama-3.2\\ Vision\end{tabular}} & No Context & 38.66 / 86.47 \\
 & Dataset Context & 37.54 / 86.48 \\
 & Action Context & \underline{38.66 / 83.31} \\\hdashline
GPT 4o & Action Context & \underline{38.66 / 85.23} \\
DeepSeek-VL2 & Action Context & 38.18 / 82.99 \\\midrule
\multirow{3}{*}{AH$_{GT}$} & Description & 57.18 / 90.28 \\
 & Concat & \textbf{61.83 / 89.64} \\
 & Transformer & 52.18 / 87.43 \\\hdashline
AH$_{pred}$ & Concat &  \cellcolor{green!15}44.66 / 82.87 \\\bottomrule
\end{tabular}
\caption{Ablation on Action History Sources on IKEA-ASM. Underlined \& green cells denote block-best and single-frame-best under realistic settings, respectively.}

\label{tab:action-history}
\end{minipage}
\vspace{-0.4cm}
\end{table*}

Although AAG is competitive overall, it is outperformed by VLMAH and AVT on some benchmarks, though at a higher computational cost. In contrast, AAG offers notable efficiency gains by operating on a single frame, which reduces both feature extraction costs and the number of trainable parameters. Its design allows each frame to be processed once, with extracted features reused across components (e.g., the action recognition model). Moreover, AAG eliminates the necessity for a text encoder during inference by pre-encoding action classes. We compare computational efficiency in \Cref{tab:compute_comparison} with video-based methods. In comparison to AAG, AVT and TempAgg require a substantially larger number of trainable parameters. Despite its comparable model size, VLMAH relies on external action recognizers , which increases pipeline complexity and training cost due to fine-tuning, whereas AAG relies on self-supervised, task-agnostic feature extractors. RULSTM is more compact but underperforms relative to AAG. Finally, at inference time, AAG further improves efficiency by reducing memory consumption and processing shorter sequences.

\subsection{Dataset Characteristics \& Past Action History Memory Buffer Size}

One key observation is that while incorporating past context through a past action history is advantageous for single-frame action anticipation, the optimal size of the action history buffer varies significantly across datasets, highlighting their distinct structural characteristics. As presented in Table~\ref{tab:past_act_history} for IKEA-ASM, the best performance is observed when considering 7 or 3 past actions, whereas for Meccano and Assembly101, the optimal buffer size is 3. This variation is attributed to the nature of each dataset: IKEA-ASM consists of structured assembly tasks with relatively predictable action sequences, where a longer history (up to 7 actions) provides beneficial contextual cues. In contrast, Meccano and Assembly101 feature more complex interactions, where excessive past actions may introduce noise rather than useful predictive information.

\subsection{Ablation on Past Action History Sources}
Beyond action history length, we examine past action sources in Table~\ref{tab:action-history} using IKEA-ASM dataset. We investigate two approaches: (a) how a Vision-Language Model, given a single frame, can output the $N=5$ past actions, with and without context regarding the current observed action; and (b) defining the action history simply as a sequence of action labels. For both approaches, the generated sentence is forwarded to DistilBERT to extract the embedding, as described in \cref{sec:method}. Additionally, for approach (b), we explore three methods for defining the embedding: (1) feeding a single description of past actions with DistilBERT, (2) feeding each past action label to DistilBERT separately and concatenating the resulting embeddings, and (3) using a self-attention transformer encoder to aggregate each past action embedding.

Regarding the first strategy, our findings (\textit{rows 1-5}) indicate that even state-of-the-art VLMs (GPT-4o, LLama-3.2 Vision, and DeepSeek-VL2), are less accurate in generating informative past action history when conditioned on a single frame, even with task- and dataset-specific context, compared to the strategy of generating the action history by working directly with the action labels that have been generated through a single-frame action recognizer (\textit{rows 6-8} oracle, \textit{row 9} realistic). This suggests that while VLMs possess strong generalization capabilities, they may lack the fine-grained temporal and contextual understanding required to infer past actions reliably from a static visual cue.


Overall the experimental results highlight that the strategy of generating the action history directly from the action labels produced by the action recognizer leads to a more accurate and informative representation of past actions. Across all examined schemes for defining the text embedding of the history under this strategy ((1)-(3) in \textit{rows 6-8}), the best results were achieved with method (2), where each past action label was individually processed by DistilBERT before concatenating the resulting embeddings. This method likely benefits from preserving the distinct semantic representation of each action while still capturing their temporal relationships through concatenation. Notably, this approach yields better results than the VLMs under realistic settings (AH$_{prd}$ in \textit{row 9}).



\section{Conclusions}
\label{sec:conclusions}
We demonstrated that single-frame action anticipation, when fused with the right modalities, can achieve competitive performance compared to video-based approaches while remaining significantly more computationally efficient. However, its effectiveness strongly depends on dataset characteristics such as activity complexity and scenario variability.
Among the modalities, depth was found to add value to the model, though its effect diminishes in close-up scenarios. In contrast, a key finding is that long-term context in the form of action history, significantly boosts performance, contingent on accurate action recognition. By reusing the feature representations generated for anticipation within a recognition setup as implemented in AAG, we obtain competitive recognizers without significantly increasing the overall computational cost. To further improve single-frame anticipation, integrating dataset-specific visual cues and incorporating a memory mechanism for past actions may prove essential. These findings suggest that with the right design choices, single-frame models can offer a viable and efficient alternative to more computationally demanding temporal architectures.


\section*{Acknowledgments}
This work has been funded by the Valencian regional government CIAICO/2022/132 Consolidated group project AI4Health, the Spanish State Research Agency (AEI) and ERDF/EU under grant: GEMELIA PID2024-161711OB-I00. This work has also been supported by a Spanish national fund for PhD studies, (FPU21/00414) as well as by the European Union (EU - HE Magician – Grant Agreement 101120731).

{
    \small
    \bibliographystyle{ieeenat_fullname}
    \bibliography{main}
}

\renewcommand*{\thesection}{\Alph{section}}
\renewcommand*{\thefigure}{\Alph{figure}}
\renewcommand*{\thetable}{\Alph{table}}
\setcounter{section}{0}
\setcounter{table}{0}
\setcounter{figure}{0}
\clearpage
\setcounter{page}{1}
\maketitlesupplementary
\appendix
\pagenumbering{roman}

The supplementary materials are organized as follows: \Cref{sec:suppl_ablation} presents an extended ablation study, including the effect of the action recognizer on action history performance, an analysis of visual and textual features, and fusion strategies for visual and multimodal inputs in AAG. This section also includes qualitative results on depth frames per dataset. Additionally, we provide an ablation on the effect of modalities using video inputs.
Finally, we present qualitative results on the performance of VLMs and VQA models for generating action history in \Cref{sec:suppl_vlms}.

\section{Extended Ablation}\label{sec:suppl_ablation}
\subsection{Effect of the Performance of the Single-Frame Action Recognition Model}
As outlined in the main paper, the action history performs best when it is generated using action predictions provided by an action recognition method. In all experiments, a single-frame action recognition model with the same architecture as the proposed AAG model was employed, effectively adapting AAG for action recognition. As shown in Table~\ref{tab:single-framea-r}, the performance of this module is generally lower than video-based approaches, including those using temporal aggregation with AAG, as well as state-of-the-art (SOTA) methods. However, it is worth noting that single-frame AAG outperforms SOTA methods on the IKEA-ASM dataset.

Given that the accuracy of this module can influence the overall performance of the proposed AAG method, as demonstrated by baseline results with ground-truth action history, we conducted an ablation study to evaluate how the performance of the action recognition model impacts the quality of past action history and, consequently, action anticipation accuracy.

Table \ref{tab:ar-extended} shows the performance of AAG across the three benchmarks. In each case, we examined the impact of using (a) the single-frame action recognition method, and (b) the best performing video-based action recognition model reported in \Cref{tab:single-framea-r}, to populate the past action history.

By enhancing the model's ability to recognize actions, we observe a significant improvement in the overall performance of the proposed single-frame action anticipation method. This improvement can be attributed to the higher-quality action recognition model, which generates more accurate and reliable action history, directly influencing the performance of the action anticipation task. When past actions are correctly predicted, they provide better context for anticipating the next action, improving accuracy and reducing ambiguity in the current scene, which ultimately enables the model to make more accurate predictions. While the observed improvements are consistent across the three benchmarks, this effect is particularly pronounced on the Meccano dataset, in which AAG previously underperformed when action history was based on predictions. The lower performance of the action recognizer is partly attributed to the challenges of temporal aggregation, particularly in the presence of rapid view changes in head-mounted videos. However, we argue that the multimodal approach as in \cite{RAGUSA2023103764}, incorporating both object and hand information in the egocentric perspective, could provide valuable insights for future work on single-frame action anticipation. 

Finally, it is worth noting that the modality contribution per dataset is consistent with previous results, demonstrating that the action history improves the robustness of the results, while not disturbing the conclusions drawn from previous analysis.

\begin{table}[t]
    \centering
    \footnotesize
    \setlength{\tabcolsep}{3pt} 

\resizebox{\linewidth}{!}{%
    \begin{tabular}{lccc}
    \toprule
    \textbf{Methods} & \textbf{IKEA-ASM} & \textbf{Meccano} & \textbf{Assembly101} \\ \midrule
    Video (RGB-only) & 57.58 / - & 45.16 / 73.75 & - \\
    Video (Multimodal) & 64.25 / - & \textbf{49.66 / 73.75} & \textbf{43.60} / - \\\midrule
    AAG$_{AR}$ (RGB + AH) & 66.43 / 92.56 & 31.21 / 66.03 & 34.19 / 58.30 \\
    AAG$_{AR}$ (RGB + Depth + AH) & 66.84 / 92.29 & 31.14 / 66.52 & 32.77 / 56.19 \\ \hdashline
    AAG$_{AR-vid}$ (RGB + AH) & 70.54 / 93.17 & 32.38 / 66.06 & 34.73 / 60.20 \\
    AAG$_{AR-vid}$ (RGB + Depth + AH) & \textbf{71.35 / 94.96} & 33.48 / 69.32 & 34.87 / 60.77 \\ 
    \bottomrule
    \end{tabular}
    }
    \caption{Action recognition performance report (Top-1/5) of single-frame action recognition with the same structure as AAG. Baselines include: IKEA-ASM uses I3D~\cite{carreira2017quo} (Top-5 is not provided); Meccano uses SlowFast~\cite{Feichtenhofer_2019_ICCV}; Assembly101 uses TSM~\cite{lin2019tsm} and is trained on all fixed cameras.}
    \label{tab:single-framea-r}
\end{table}

\begin{table}[t]
    \centering
    \footnotesize
    \setlength{\tabcolsep}{3pt} 
    \resizebox{\linewidth}{!}{%
    \begin{tabular}{l l cc cc c}
        \toprule
        \textbf{AH Source} & \textbf{Modalities} & \multicolumn{2}{c}{\textbf{IKEA-ASM}} & \multicolumn{2}{c}{\textbf{Meccano}} & \textbf{Assembly101} \\
        \cmidrule(lr){3-4} \cmidrule(lr){5-6} \cmidrule(lr){7-7}
        & & Top-1 & Top-5 & Top-1 & Top-5 & Recall@5 \\
        \midrule
        \multirow{3}{*}{Single-Frame} 
            & AH* & 41.42 & 79.03 & 25.72 & 51.97 & \underline{9.80} \\
            & RGB + AH & 43.82 & 83.11 & 26.22 & 52.18 & 8.25 \\
            & RGB + Depth + AH & \underline{44.66} & \underline{82.87} & \underline{26.43} & \underline{54.95} & 8.90 \\
        \midrule
        \multirow{3}{*}{Best Video AR} 
            & AH* & 44.50 & 82.31 & 27.78 & 59.17 & \textbf{15.93} \\
            & RGB + AH & 47.46 & 84.51 & 27.46 & 52.05 & 11.23 \\
            & RGB + Depth + AH & \textbf{48.02} & \textbf{85.03} & \textbf{28.27} & \textbf{58.46} & 11.35 \\
        \bottomrule
    \end{tabular}}
    \caption{Single-frame anticipation results under single-frame and video-based action histories. Underline indicates best single-frame. * AH generated from RGB, Depth, AH action recognizer, while the other AH is generated from the same inputs as those used in the anticipation-based AAG.}
    \label{tab:ar-extended}
\end{table}

\begin{figure*}[t]
    \centering
    \includegraphics[width=0.9\linewidth]{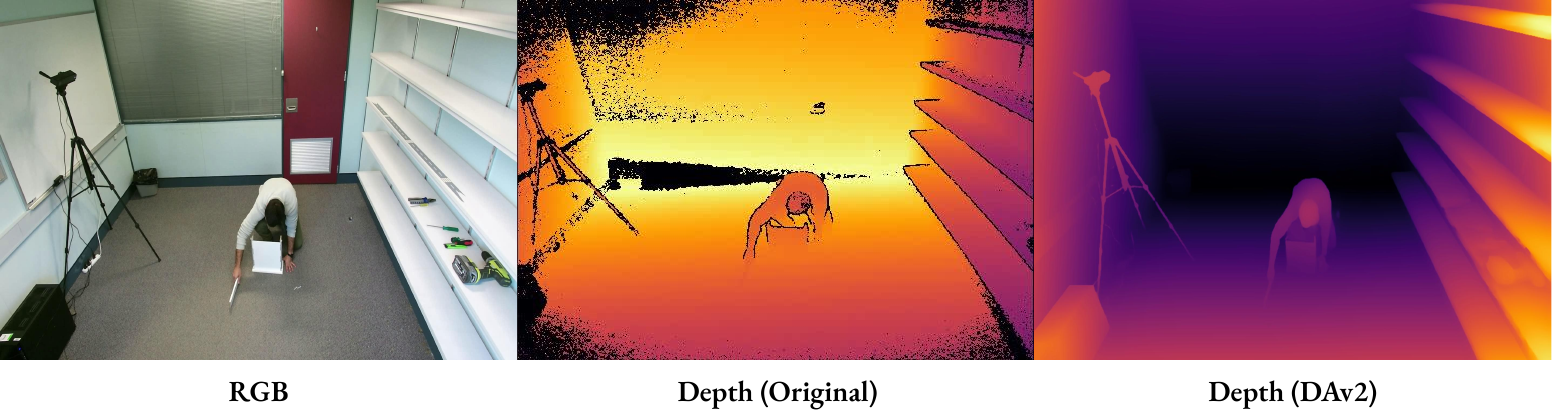}
    \caption{Comparison of original depth frames and Depth Anything v2 estimates on the IKEA-ASM dataset. The higher noise in the original depth frames causes reduced performance when fused with RGB features.}
    \label{fig:depth}
\end{figure*}

\subsection{Analysis on Visual Modalities}
In the main paper, we show that depth information is more useful when extracted using Depth Anything V2 (DAv2)~\cite{NEURIPS2024_26cfdcd8}. Similarly, we find that DINOv2~\cite{oquab2023dinov2} combined with a cross-attention fusion mechanism is most effective for obtaining visual representations. This section ablates different strategies and encoders to validate these findings on the IKEA-ASM dataset.

\noindent \textbf{Depth}: is considered an auxiliary modality that contributes to a more robust visual representation when fused with RGB data. Choosing an appropriate fusion strategy is key to obtain optimal results. This statement is supported by the results in \Cref{tab:ikea-rgb_depth-fusion}. This table ablates different fusion strategies: concatenating and projecting the embeddings; sum of embeddings; soft-attention, defined as a softmax operation on depth embeddings followed by a dot product; self-attention using a transformer encoder; and cross-attention under the two possible settings, RGB or depth as query and the respective as key/value. Additionally, this table shows that depth information from ground truth, despite being more effective individually than the estimate, it shows diminished performance when fused with RGB data. As remarked in Section~\ref{sec:method}, we attribute this to the noisier nature of the depth ground-truth images,  and shown in \Cref{fig:depth}, where depth images compared to DAv2 estimates are noisier and may disrupt the fusion process, leading to suboptimal results.

Additionally, we can observe that our experiments revealed that among various fusion techniques for combining RGB and Depth features, cross-attention with $Q:$ RGB achieved the best performance. Our intuition is that this approach allows rich semantic and contextual cues from RGB data to guide the attention mechanism in selecting relevant information from the depth modality. Essentially, by leveraging RGB as queries, the model effectively prioritizes spatial and geometric features from the depth map that align with the observed scene, leading to more accurate and context-aware action anticipation. 

\begin{table}[t]
\centering
\footnotesize
\setlength{\tabcolsep}{3pt} 
\begin{tabular}{lcr}
\toprule
\multicolumn{1}{c}{\textbf{Fusion Strategy}} & \textbf{Depth Source} & \textbf{Top-1 / 5 Acc} \\\midrule
Depth-Only & \multirow{7}{*}{GT} & 28.25 / 74.55 \\
Concatenation & & 27.73 / 69.95 \\
Sum &  & 29.25 / 73.19 \\
Soft-Attention &  & 27.69 / 74.79 \\
Self-Attention &  & 37.09 / 85.39 \\
Cross-Att Q: Depth & & 26.65 / 71.91 \\
Cross-Att Q: RGB & & \underline{37.29 / 85.23} \\\midrule
Depth-Only & \multirow{7}{*}{DAv2} & 26.33 / 68.95 \\
Concatenation &  & 35.53 / 82.43 \\
Sum &  & 33.85 / 78.07 \\
Soft-Attention &  & 25.85 / 73.71 \\
Self-Attention &  & 38.34 / 85.23 \\
Cross-Att Q: Depth & & 26.33 / 68.95 \\
Cross-Att Q: RGB &  & \textbf{38.82 / 86.19} \\\bottomrule
\end{tabular}
\caption{Ablation on Visual Fusion Strategies and Depth Source.}
\label{tab:ikea-rgb_depth-fusion}
\end{table}

\begin{figure}[ht]
    \centering
    \includegraphics[width=\linewidth]{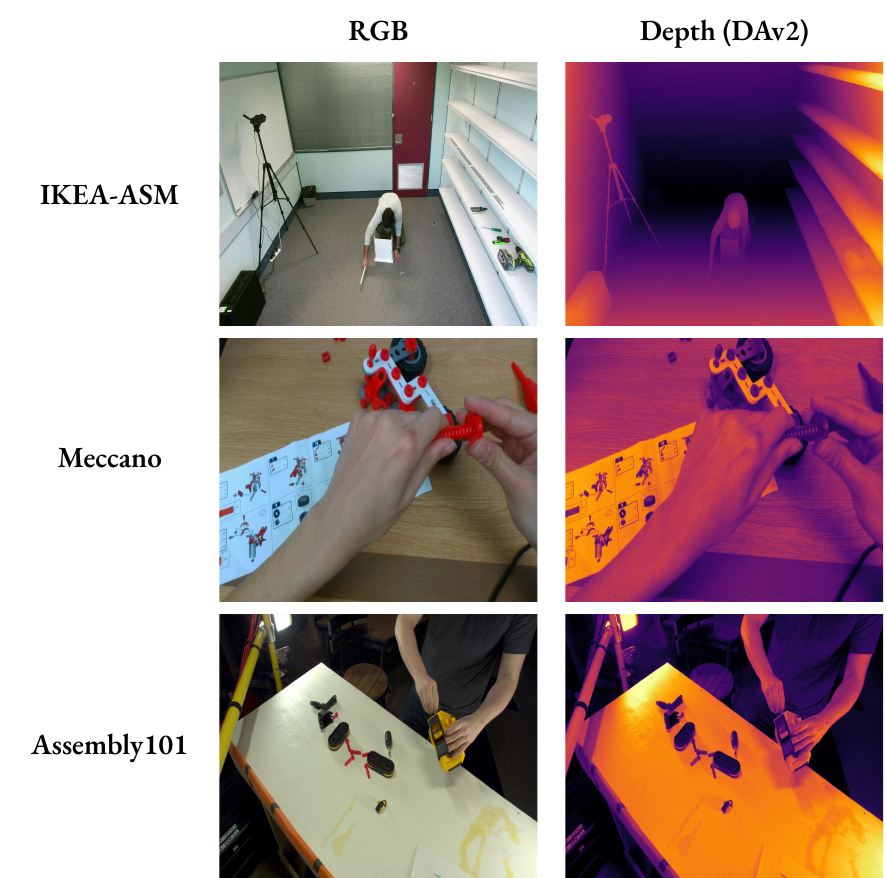}
    \caption{Qualitative comparison of depth frames from DAv2 against the original RGB images across different benchmark datasets. The third-person perspective in IKEA-ASM clearly distinguishes background and foreground elements, enhancing depth's utility. In contrast, the closer camera views in Meccano and Assembly101, where most objects and interactions occur within a nearly planar workspace, limit the effectiveness of depth.}
    \label{fig:depth-ablation}
\end{figure}

Finally, we provide some qualitative results on depth frames across the selected benchmark datasets to support the statements in Section~\ref{sec:results} regarding the contribution of depth across camera views. Figure~\ref{fig:depth-ablation} provides samples of depth frames from DAv2 compared to the original RGB image. While the third-person perspective on the IKEA-ASM dataset allows for clear distinction between background and foreground elements, the closer camera view in Meccano and Assembly101 limits the effectiveness of depth in these close-up perspectives. As a result, depth fails to offer valuable additional information and, in some cases, can even be misleading in these datasets, as shown in the experimental results. This further reinforces the varying utility of depth depending on the camera perspective, as discussed in Section~\ref{sec:results}.

\noindent \textbf{Backbone Feature Extractor:} The choice of an adequate feature extractors for RGB and depth features has a considerable impact on the performance of AAG. \Cref{tab:ikea-rgb_feats} ablates 3 different features extractors, (a) the 2D-CNN BN-Inception pre-trained on the TSN action recognition architecture~\cite{wang2018temporal}, (b) ViT-L-14~\cite{dosovitskiy2020image} pre-trained on CLIP~\cite{radford2021learning}, and (c) DINOv2~\cite{oquab2023dinov2}. From the three, the best encoding performance on both RGB and depth modalities is given by DINOv2, whose self-supervised learning, enables us to derive richer and more generalized representations by removing the need for explicit labels during training.

\begin{table}[ht]
\footnotesize
\setlength{\tabcolsep}{3pt}
\centering
\begin{tabular}{lccr}
\toprule
\textbf{Modality} & \textbf{Feature Extractor} & \textbf{Top-1 / 5 Acc} \\\midrule
\multirow{3}{*}{RGB} & DINOv2 & \textbf{34.37 / 83.43} \\
 & ViT-L-14 & 26.33 / 69.07 \\
 & TSN BN Inception & 26.33 / 68.95 \\\midrule
\multirow{2}{*}{Depth DAv2} & DINOv2 & 26.33 / 68.95 \\
 & ViT & 26.33 / 69.07 \\ \bottomrule
\end{tabular}
\caption{Ablation on Visual Feature Extractors.}
\label{tab:ikea-rgb_feats}
\end{table}

\subsection{Analysis on Multimodal Fusion}
As for the visual fusion, we conduct an ablation on multimodal fusion strategies and text encoders under the best settings of AAG: RGB, depth and AH$_{gt}$, using a concatenation of the independent class name embeddings. Similarly, we use IKEA-ASM dataset. \Cref{tab:text-encoder} depicts five different text encoders for AAG. Not only DistilBERT provides the best performance. We also evaluate CLIP's text encoder on DINOv2 and CLIP's ViT features. We find that the alignment between textual and visual embeddings for this task yields the worst results, showcasing the ability of DINOv2 even when compared to visual and textual features in the same embedding space.

\begin{table}[ht]\centering
\footnotesize
\setlength{\tabcolsep}{3pt}
\centering
\begin{tabular}{lr}
\toprule
\textbf{Text Encoder} & \textbf{Top-1 / 5 Acc} \\\midrule
BERT \cite{devlin2019bert} & 60.42 / 90.20 \\
RoBERTa \cite{liu2019roberta} & 56.78 / 88.68 \\
DistilBERT \cite{sanh2019distilbert} & \textbf{61.83 / 89.64} \\
CLIP Text Encoder \cite{radford2021learning} & 39.66 / 84.83 \\
CLIP Text + Visual Encoder \cite{radford2021learning} & 27.89 / 68.95 \\\bottomrule
\end{tabular}
\caption{Ablation on text encoder.}
\label{tab:text-encoder}
\end{table}

\Cref{tab:ikea-visual_text-fusion} ablates four different multimodal fusion strategies: concatenation of visual and textual features; sum of embeddings; a single self-attention transformer to fuse the three modalities, \emph{i.e.,} not using cross-attention to fuse visual features first; and a self-attention transformer on visual and textual features, which is the one selected for AAG due to its superior performance.

\begin{table}[ht]
\footnotesize
\setlength{\tabcolsep}{3pt}
\centering
\begin{tabular}{lr}
\toprule
\multicolumn{1}{c}{\textbf{Fusion Strategy}} & \textbf{Top-1 / 5 Acc} \\ \midrule
Concat (Visual CA + Text) & 56.06 / 87.23 \\
Sum (Visual CA + Text) & 51.22 / 86.43 \\
Self-Attention (RGB + Depth + Text) & 58.38 / 88.96 \\
Self-Attention (Visual CA + Text) & \textbf{61.83 / 89.64} \\ \bottomrule
\end{tabular}
\caption{Ablation on Visual-Text Fusion Strategies. CA = Cross-Attention.}
\label{tab:ikea-visual_text-fusion}
\end{table}

\subsection{Impact of Multimodal Cues Under Temporal Aggregation}
We extend the analysis presented in \Cref{sec:results} and particularly \Cref{tab:modalities} to provide further insights on the effect of the selected modalities under temporal aggregation of video frames. First we compare the temporal aggregation mechanism from prior work~\cite{benaventlledo2024enhancingactionrecognitionleveraging,Wang_2021_ICCV} with self-supervised video transformers. The temporal aggregation method consists of a transformer encoder with 3 layers and 8 attention heads per layer, fixed positional encoding to preserve temporal order, and a CLS token as a global representation. This strategy is selected as a simple baseline approach to temporal aggregation, having demonstrated superior performance compared to prior approaches, such as RNNs, in capturing temporal dependencies.

\begin{table}[ht]
\footnotesize
\centering
\setlength{\tabcolsep}{3pt} 
\begin{tabular}{lrr}
\toprule
\textbf{Method} & \textbf{Top-1} & \textbf{Top-5} \\\midrule
Video Transformer Encoder & \textbf{63.15} & \textbf{91.76} \\
V-JEPA 2~\cite{assran2025v} & 58.58 & 90.96 \\
TimeSformer~\cite{bertasius2021space} & 57.22 & 90.24 \\\bottomrule
\end{tabular}
\caption{Comparison of video feature extractors and temporal aggregation methods on AAG. Inputs: RGB, Depth, AH$_{GT}$.}
\label{tab:video-enc}
\end{table}

\begin{table*}[t]
\footnotesize
\setlength{\tabcolsep}{3pt}
\centering
\begin{tabular}{l rr rr rr}
\toprule
\textbf{Modalities} & \multicolumn{2}{c}{\textbf{IKEA-ASM}} & \multicolumn{2}{c}{\textbf{Meccano}} & \multicolumn{2}{c}{\textbf{Assembly101}} \\\cmidrule(lr){2-3} \cmidrule(lr){4-5} \cmidrule(lr){6-7}
 & \textbf{Top1 / 5 Acc} & \textbf{Recall@5} & \textbf{Top1 / 5 Acc} & \textbf{Recall@5} & \textbf{Top1 / 5} & \textbf{Recall@5} \\\midrule

RGB & 36.61 / 86.47 & 44.50 & 27.21 / 49.52 & 8.33 & 6.41 / 21.53 & 4.19 \\
RGB, Depth & 37.58 / 85.15 & 46.35 & 27.21 / 52.11 & 8.33 & 6.15 / 22.66 & 3.63 \\\midrule
AH$_{GT}$ & \textbf{63.55 / 94.52} & 65.09 & 31.75 / 73.57 & 37.33 & 31.19 / 59.70 & \textbf{38.49} \\
AH$_{Pred}$* & 44.50 / 82.31 & 55.88 & 27.60 / 52.89 & 13.01 & 13.99 / 34.60 & 12.55 \\\midrule
RGB, AH$_{GT}$ & 62.46 / 90.12 & 55.36 & 30.29 / 63.75 & 17.46 & 29.32 / 56.15 & 32.14 \\
RGB, AH$_{Pred}$ & 41.06 / 82.07 & 45.94 & 26.43 / 55.52 & 13.25 & 14.06 / 31.77 & 10.08 \\\midrule
RGB, Depth, AH$_{GT}$ & 63.15 / 91.76 & 60.97 & \textbf{32.88 / 67.90} & 25.95 & 30.27 / 56.51 & 33.05 \\
RGB, Depth, AH$_{Pred}$ &\underline{ 46.02 / 83.43} & 50.79 & \underline{27.24 / 54.52} & 12.53 & 14.01 / 31.97 & \underline{11.24} \\\bottomrule
\end{tabular}
\caption{Analysis of the influence of modality selection on AAG performance with video aggregation performed on visual inputs. Underlined values indicate the best performance under realistic settings. * denotes that the action history was generated using the RGB, Depth, AH action recognizer. Otherwise, AH is generated from the same inputs as those used for anticipation.}
\label{tab:video-modalities}
\end{table*}

By leveraging the same visual inputs as the single-frame method, we can evaluate the effect of temporal aggregation under the same training conditions. While the use of this training schedule may limit the performance of the video transformer, it is important to note that optimizing the video transformer is not the main goal of this work. Instead, the objective is to quantify its performance differences under identical conditions. On the other hand, self-supervised video transformers enable a one-to-one comparison between image-based and video-based self-supervised models, providing a clearer understanding of how temporal aggregation influences performance across modalities. Results in \Cref{tab:video-enc} show that self-supervised temporal aggregation fails to provide competitive performance, including both TimeSformer~\cite{bertasius2021space} and V-JEPA 2~\cite{assran2025v}. We attribute this to the fact that aggregating self-supervised image features leverages robust individual frame representations, which are highly optimized for the task, while self-supervised video transformers may struggle to capture fine-grained temporal dependencies without fine-tuning on a specific task.

\Cref{tab:video-modalities} presents an extended ablation of \Cref{tab:modalities}, incorporating video aggregation on the visual modalities. The results support our previous single-frame analysis. On visual inputs, we observe an improvement in IKEA-ASM when including depth, though this effect is diminished and only apparent in Top-5 accuracy for Meccano. As with the single-frame approach, depth is misleading when fused with RGB information in Assembly101. Action history, when populated from ground-truth data, continues to deliver the strongest performance on both IKEA-ASM and Assembly101. Similar to the single-frame approach, visual inputs yield misleading results compared to action history when it is generated from predictions of an equivalent action recognizer on Assembly101. For detailed accuracy results of each action recognizer, refer to \Cref{tab:single-framea-r}.

Given the results of state-of-the-art video-based methods on these datasets, we attribute the lower performance observed here to the relatively simple transformer encoder used as a baseline. More refined temporal modeling, along with longer windows as seen in existing state-of-the-art approaches, is necessary to effectively capture interactions in datasets like Assembly101, which feature long-tail action distributions and high variability in the performed actions.

Notably, results on the Meccano dataset effectively leverage the three modalities. Despite the lower contribution of depth when combined with RGB alone, it still produces the strongest result in this setting. Additionally, the improvement in action recognition accuracy, along with the temporal aggregation of short-term observations, leads to the most effective results when all three modalities are fused, as highlighted in our single-frame analysis.


\section{VLM Performance: Qualitative Evaluation} \label{sec:suppl_vlms}

\begin{figure*}[p]
    \centering
    \includegraphics[width=\textwidth]{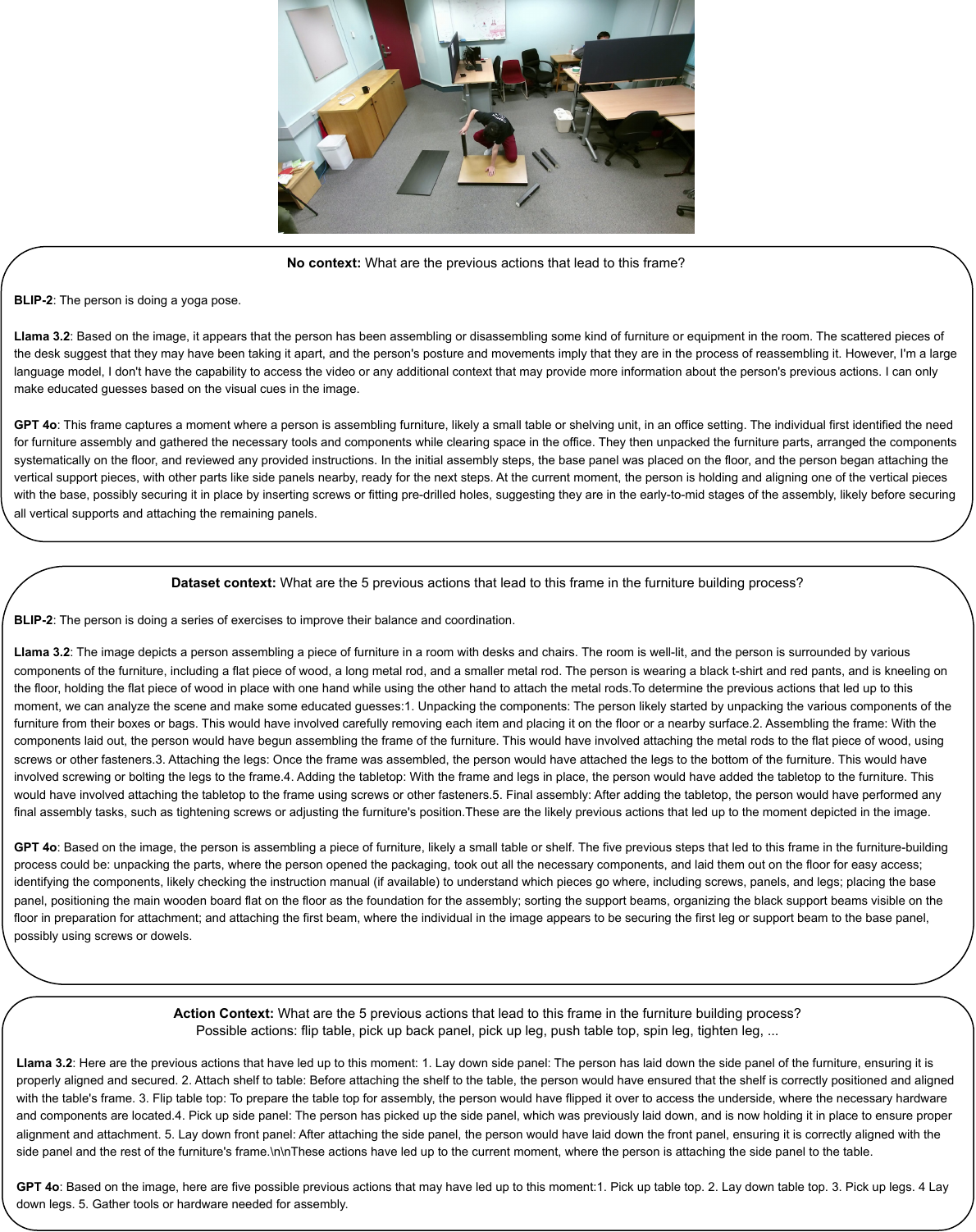}
    \caption{Qualitative comparison of VLMs and VQA methods on 3 different prompts. Frame extracted from IKEA-ASM.}
    \label{fig:vlms-1}
\end{figure*}

This section presents a qualitative analysis of vision-language models and extends the evaluation to the state-of-the-art visual question-answering (VQA) model, BLIP-2~\cite{li2023blip}. \Cref{fig:vlms-1} provides qualitative comparisons of responses generated by Llama-3.2 Vision, GPT-4o, and BLIP-2 based on selected prompts.

BLIP-2 demonstrates a lack of reasoning beyond visual features, primarily focusing on human pose while disregarding contextual cues in the prompt. In contrast, VLMs exhibit a more human-like reasoning process, though they still present notable limitations, as acknowledged by the models themselves. For instance, Llama-3.2 (without context) explicitly states its difficulty in analyzing a single frame without additional information: \emph{``...However, I'm a large language model, I don't have the capability to access the video or any additional context that may provide more information about the person's previous actions. I can only make educated guesses based on the visual cues in the image''}.

When contextual information is provided, VLM responses become more detailed and scenario-driven. Including domain-specific knowledge, such as action class labels, leads to more concise and informative outputs, particularly in GPT-4o, which demonstrates stronger reasoning capabilities in structured environments.

\begin{figure*}[htbp]
    \centering
    \includegraphics[width=\textwidth]{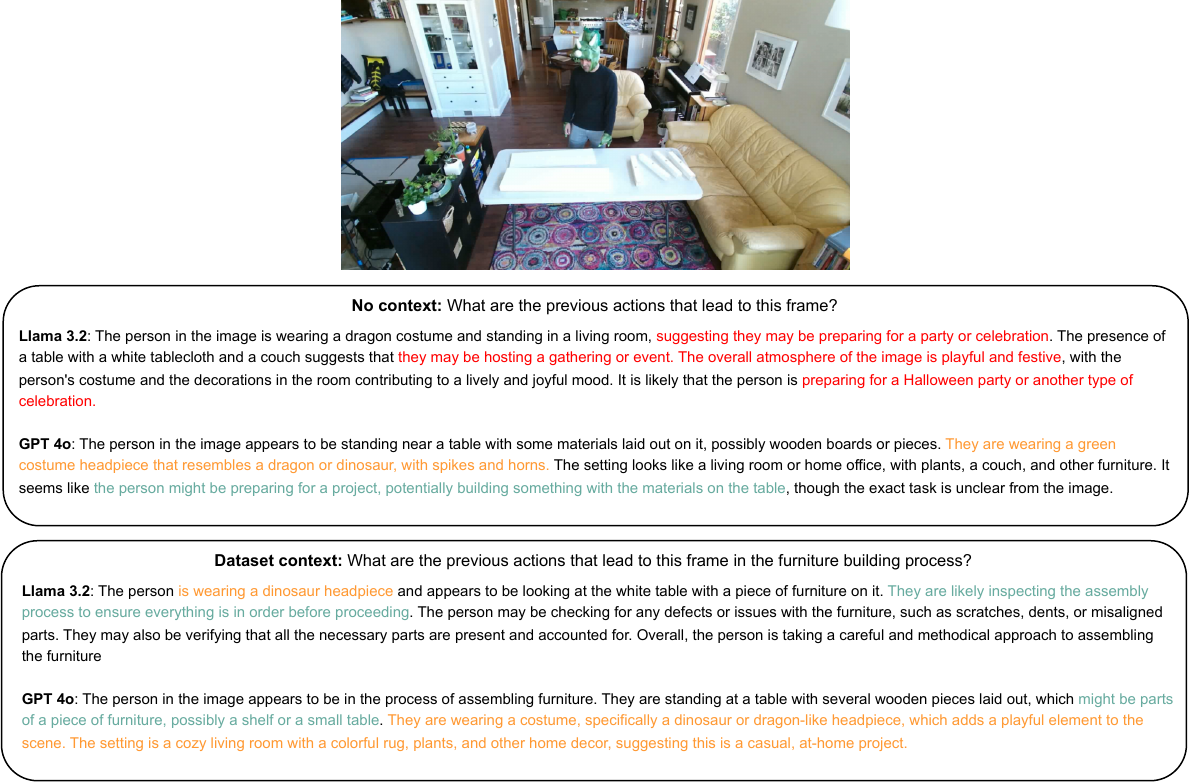}
    \caption{Qualitative results on the importance of contextual prompting. Frame extracted from IKEA-ASM. We use red color to highlight action-related errors, green for action-related valid answers, and orange for non-related visual descriptions.}
    \label{fig:vlms-2}
\end{figure*}

The relevance of contextual information in the prompts is further demonstrated in \Cref{fig:vlms-2}. We include 2 prompts with and without dataset context. In this prompts we do not include a specific number of actions either, showcasing the importance of such information to get more structured responses. The main issues we may find in this descriptions is that due to the lack of context, Llama 3.2 focuses on a single item, \emph{a dinosaur headpiece} and its responses is celebration-oriented (\emph{``...suggesting they may be preparing for a party or a celebration...''}) rather than assembly-oriented. GPT 4o in contrast, while acknowledges the item is able to recognize the furniture pieces, though without specifying any particular previous task, which in this case, would be even hard for humans to do. By providing context, Llama 3.2 offers a more detailed description than GPT 4o, assuming that prior steps include preparation and inspection of the parts. This frame in particular corresponds to the beginning of a video sequence, so there are no prior actions annotated and the estimate provided by Llama is accurate enough. Notably, these prompts do not specify a fixed number of past actions, highlighting the role of structured input in obtaining more coherent responses compared to the ones in \Cref{fig:vlms-1}.

Overall, VLMs demonstrate strong general reasoning abilities but remain less effective for dataset-specific, fine-grained tasks. As a result, their generated descriptions can be misleading, whereas simple action labels provide more reliable and consistent guidance for AAG.

\end{document}